\documentclass[fleqn,10pt]{wlscirep}
\usepackage[utf8]{inputenc}
\usepackage[T1]{fontenc}
\usepackage{subcaption}
\usepackage{bbm}
\usepackage{multirow}

\title{EPEE: Towards Efficient and Effective Foundation Models in Biomedicine}

\author[1]{Zaifu Zhan}
\author[2]{Shuang Zhou}
\author[3]{Huixue Zhou}
\author[4]{Zirui Liu}
\author[2,*]{Rui Zhang}
\affil[1]{Department of Electrical and Computer Engineering, University of Minnesota, Minneapolis, MN, United States}
\affil[2]{Division of Computational Health Sciences, Department of Surgery, University of Minnesota, Minneapolis, MN, United States}
\affil[3]{Institute for Health Informatics, University of Minnesota, Minneapolis, MN 55455, United States}
\affil[4]{Department of Computer Science and Engineering, University of Minnesota, Minneapolis, MN, United States}
\affil[*]{Corresponding author: Rui Zhang (\href{zhan1386@umn.edu}{zhan1386@umn.edu})}

\begin{abstract}
Foundation models, including language models, e.g., GPT, and vision models, e.g., CLIP, have significantly advanced numerous biomedical tasks. 
Despite these advancements, the high inference latency and the ``overthinking'' issues in model inference impair the efficiency and effectiveness of foundation models, thus limiting their application in real-time clinical settings.
To address these challenges, we proposed EPEE (Entropy- and Patience-based Early Exiting), a novel hybrid strategy designed to improve the inference efficiency of foundation models. 
The core idea was to leverage the strengths of entropy-based and patience-based early exiting methods to overcome their respective weaknesses.
To evaluate EPEE, we conducted experiments on three core biomedical tasks—classification, relation extraction, and event extraction—using four foundation models (BERT, ALBERT, GPT-2, and ViT) across twelve datasets, including clinical notes and medical images. The results showed that EPEE significantly reduced inference time while maintaining or improving accuracy, demonstrating its adaptability to diverse datasets and tasks.
EPEE addressed critical barriers to deploying foundation models in healthcare by balancing efficiency and effectiveness. It potentially provided a practical solution for real-time clinical decision-making with foundation models, supporting reliable and efficient workflows.

\end{abstract}

\begin{document}

\flushbottom
\maketitle
%
%
\thispagestyle{empty}

\section{Introduction}
Foundation models, including language models, e.g., BERT~\cite{devlin-etal-2019-bert} and GPT series~\cite{radford2019language}, and vision models, e.g., Vision Transformers (ViTs)~\cite{dosovitskiy2021an} and CLIP~\cite{pmlrv139radford21a}, have become increasingly popular in artificial intelligence, setting new benchmarks across diverse tasks~\cite{azad2023foundational,lu2024visual}. Their impact is particularly significant in healthcare~\cite{zhou2024large, azad2023foundational}, where language models excel in analyzing biomedical text and electronic health records (EHRs)~\cite{Shickel2018Machine}, including clinical note classification~\cite{zhan2024towards, ramie}, complex reasoning~\cite{zhou2024interpretable}, and information extraction~\cite{zhan2025mmrag, yue2020_phicon}. Similarly, vision models have demonstrated exceptional performance in medical image analysis~\cite{chen2022recent,he2023transformers}, enabling tasks such as disease detection~\cite{zhou2023foundation}, segmentation~\cite{asgari2021deep}, and classification~\cite{zhang2024generalist}.

Despite these advancements, several challenges hinder the effective application of foundation models in healthcare. First, prediction accuracy is critical, as errors can pose risks to clinicians and patients. A major challenge is addressing ``overthinking''~\cite{chen2021don,bajpai2025survey,xie2021elbert, gao2023pf}, where deeper model layers add unnecessary complexity without improving outcomes, wasting computational resources and potentially degrading performance~\cite{zhou2020bert, zhu-etal-2021-gaml}. Second, inference efficiency is paramount, especially in urgent settings like intensive care units (ICUs), where real-time, accurate decision-making is vital~\cite{morrow2012evolution}. Delays in processing medical information can result in suboptimal treatment, prolonged hospital stays, or worse outcomes. Thus, models must provide rapid and reliable assessments to support timely clinical decisions~\cite{morrow2012evolution}. However, as models grow larger, computational inefficiency and increased latency become significant barriers, impacting real-time applications and patient care workflows~\cite{zheng2024large,zhang2022pcee, Gueziri2018Latency}.

While techniques such as network pruning~\cite{zhu2017prune,xu2020bert}, knowledge distillation~\cite{jiao2020tinybert,sun2019patient}, and weight quantization~\cite{zhang2020ternarybert,kim2021bert} have been employed to improve inference efficiency, they fail to address the overthinking issue, as all layers are still used for predictions. This limits their effectiveness in biomedical contexts.
A promising alternative is early exiting, a form of adaptive inference~\cite{scardapane2020should,zhou2020bert,teerapittayanon2016branchynet,xin2020deebert}, which introduces intermediate ``exit'' points within the model. This mechanism enables simpler cases to bypass deeper layers, significantly reducing inference time while maintaining or even improving performance. By dynamically adjusting computational depth based on input complexity, early exiting not only enhances efficiency but also mitigates overthinking. Furthermore, compared to other efficiency-enhancing techniques, early exiting offers greater flexibility, allowing adjustments to meet specific real-world performance requirements, such as latency or energy consumption. These attributes make early exiting an ideal approach to improve the inference efficiency of foundation models in healthcare.

\begin{figure}[t]
\centering
\includegraphics[width=0.8\linewidth]{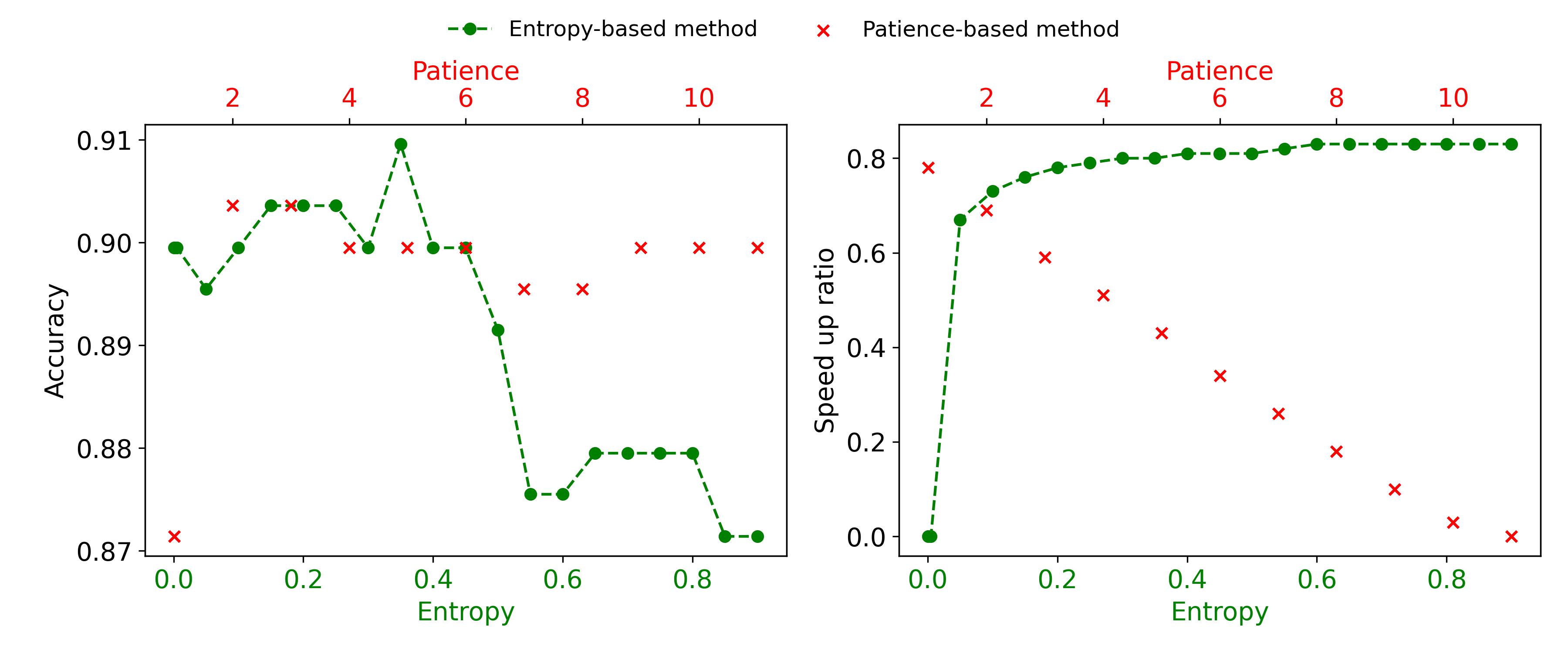}
\caption{Performance comparison of entropy-based and patience-based methods on the Dietary Supplements Usage classification task. The speed-up ratio, as defined in Section~\ref{speedup}, reflects reduced computational requirements and latency, with higher values indicating greater efficiency. 
The entropy-based method (green color) demonstrates notable efficiency and robustness at high thresholds but sacrifices accuracy.
In contrast, the patience-based method (red color) lacks robustness to the varying hyper-parameter values (i.e., patience) yet generally achieves high accuracy.}
\label{fig:prev2methods}
\end{figure}

Early exiting strategies are primarily categorized into entropy-based~\cite{kaya19a} and patience-based methods~\cite{zhou2020bert}, both of which have limitations that can lead to suboptimal performance in biomedical tasks. 
Specifically, entropy-based methods are efficient and robust under high entropy thresholds but suffer from poor accuracy (Fig.~\ref{fig:prev2methods}), limiting their reliability in clinical applications.
In contrast, patience-based methods, including PABEE~\cite{zhou2020bert}, PCEE~\cite{zhang2022pcee}, and F-PABEE~\cite{Gao2023fpabee}, allow early exiting when classifiers produce the same prediction consecutively.
While these methods generally achieve high accuracy (Fig.~\ref{fig:prev2methods}), their efficiency is highly sensitive to the hyper-parameter “patience”. Consequently, determining the optimal “patience” value to balance effectiveness and efficiency is challenging, particularly for specific clinical needs.
To address these challenges, an ideal method for clinical use should be robust to variations in its hyper-parameter, enabling easy adjustment across diverse clinical tasks while ensuring reliable and efficient decision-making~\cite{yuan2023mu, kang2015efficient}.

\begin{figure}[ht]
    \centering
    \includegraphics[width=1\linewidth]{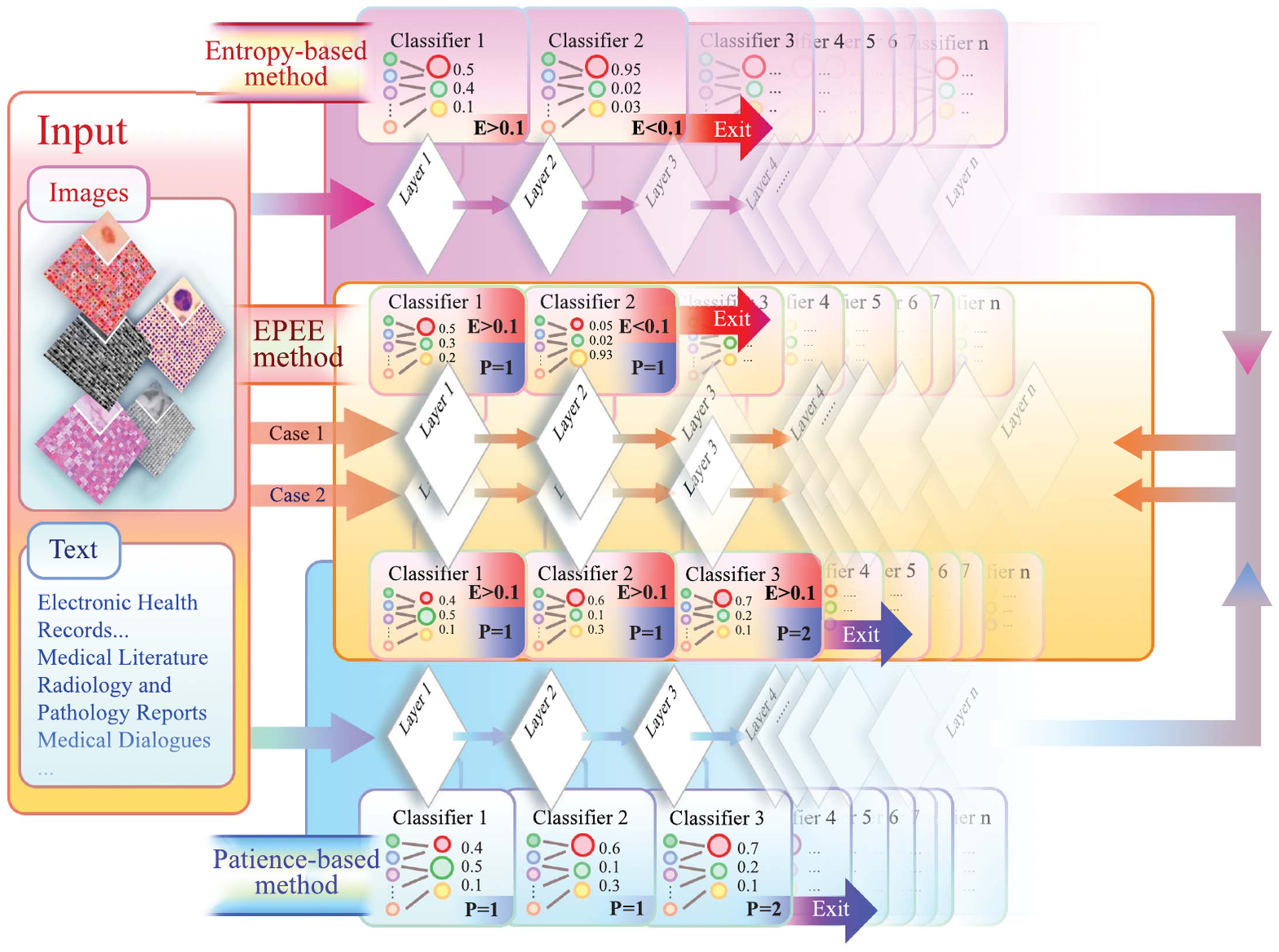}
    \caption{Overview of the proposed EPEE method.
    The method can be applied to both language and vision models. 
    The entropy-based methods exit when the entropy criterion is satisfied, and the patience-based methods exit when it reaches the pre-set patience threshold.
    Our EPEE method uses both criteria for a more general and flexible early exiting strategy.}
    \label{fig:methods}
\end{figure}

Motivated by the potential to leverage the strengths of these approaches to overcome their respective weaknesses, we proposed EPEE (Entropy- and Patience-based Early Exiting), a novel method designed to enhance the inference efficiency of foundation models, as shown in Fig. \ref{fig:methods}. EPEE incorporated intermediate classifiers at each transformer block, allowing early exits when prediction entropy was sufficiently low or predictions remained consistent across a predefined number of layers. This hybrid approach offered greater flexibility for adjusting speed-up ratios by setting both entropy and patience thresholds and was adaptable to any foundation model.
We conducted a comprehensive evaluation of EPEE on three core biomedical tasks—classification, relation extraction, and event extraction—using four commonly employed foundation models: BERT~\cite{devlin-etal-2019-bert}, ALBERT~\cite{albert}, GPT-2~\cite{radford2019language}, and ViT~\cite{dosovitskiy2021an}. Our experiments spanned eleven publicly available datasets, including clinical notes (e.g., MIMIC-ICU~\cite{hager2024evaluation}), medical images (e.g., PneumoniaMNIST~\cite{kermany2018identifying, yang2023medmnist} and PathMNIST~\cite{kather2019predicting, yang2023medmnist}), and one private dataset of clinical notes~\cite{fan2017classifying}. Results demonstrated that EPEE significantly improved inference efficiency while maintaining the effectiveness of foundation models.
Our contributions are summarized as follows:
\begin{itemize}

\item We proposed EPEE, a novel hybrid early exiting method that enhanced the inference efficiency of foundation models in biomedical applications and was applicable to any foundation model.

\item To the best of our knowledge, EPEE is the first method specifically designed to address the ``overthinking'' issue in foundation model inference, ensuring both efficiency and effectiveness.

\item Extensive experiments verified the performance of EPEE across twelve biomedical datasets, demonstrating its ability to boost the efficiency and effectiveness of four foundation models on three critical tasks.
\end{itemize}

\begin{figure}
    \centering
    \includegraphics[width=1\linewidth]{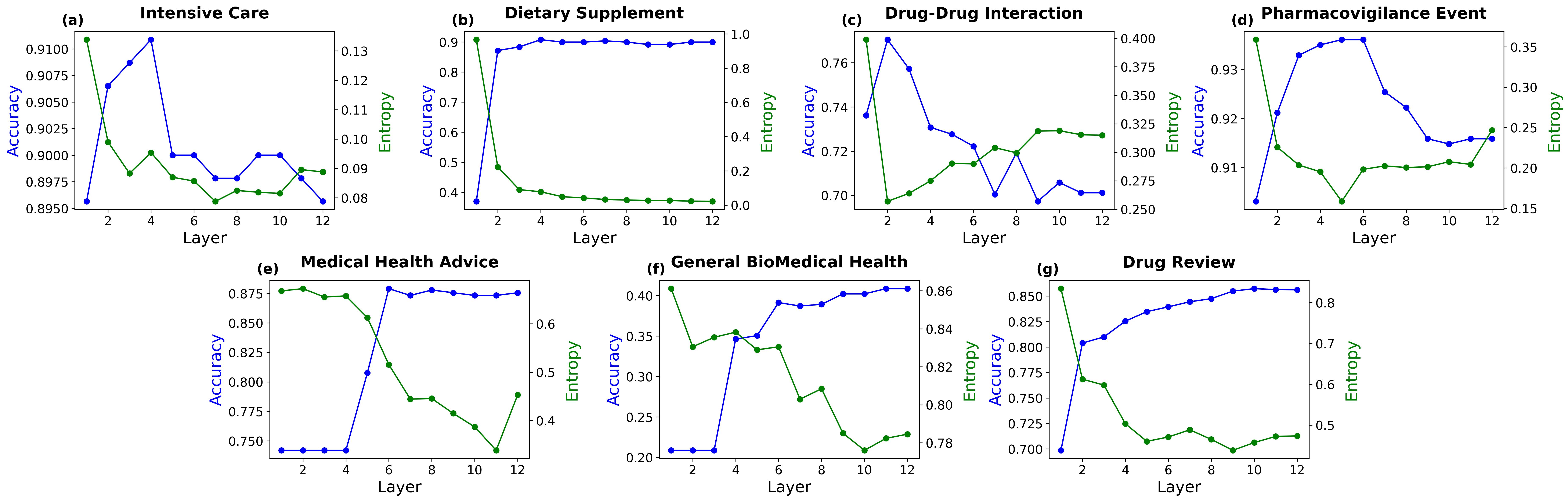}
    \caption{Entropy and accuracy across each layer of BERT in budgeted mode. The results indicate that the intermediate layer achieves performance that is comparable to or exceeds existing alternatives. Additionally, the decreasing entropy values suggest increased confidence in the predictions.}
\label{fig:effe}
\end{figure}

\begin{figure}
    \centering
    \includegraphics[width=1\linewidth]{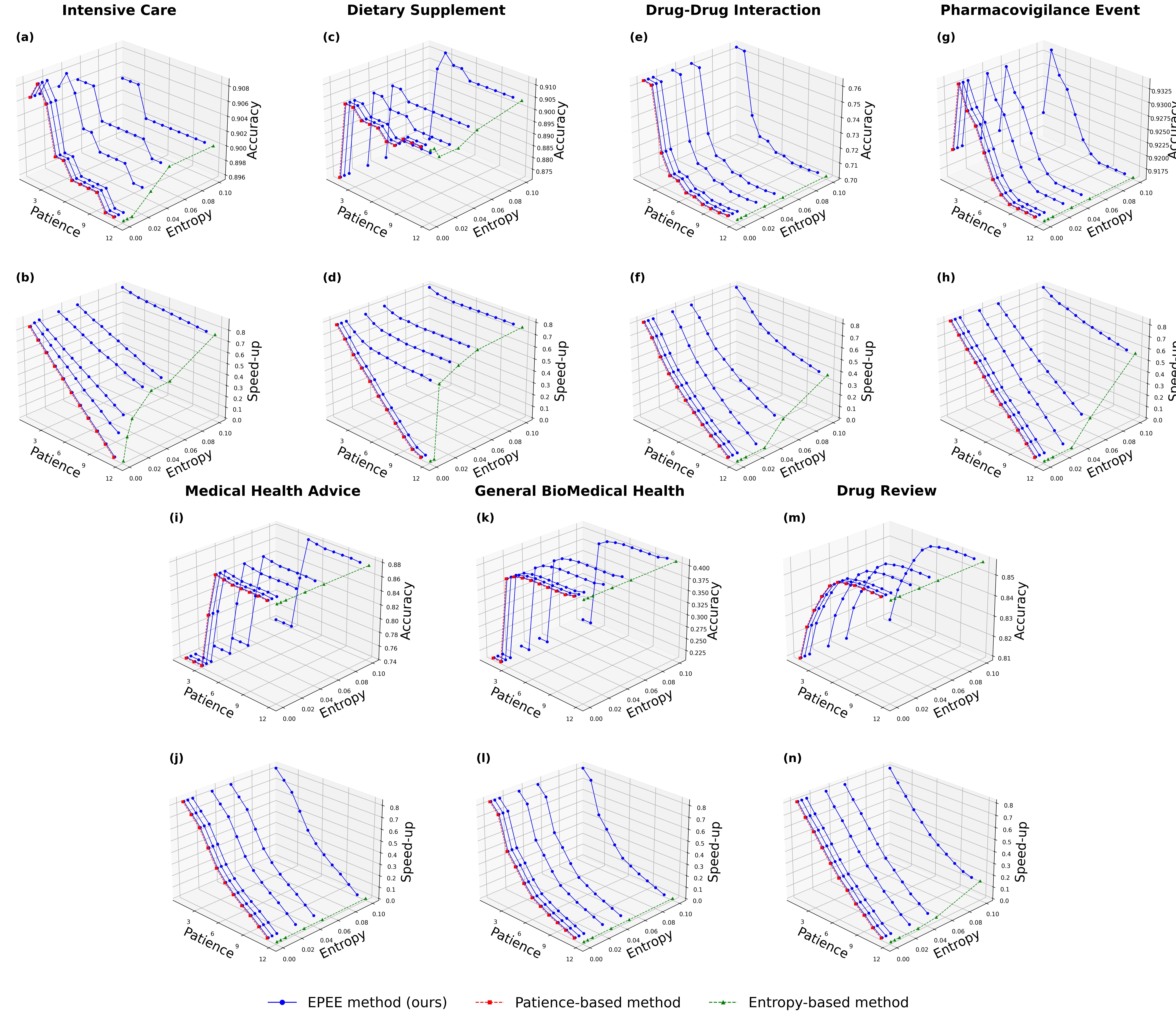}
    \caption{Dynamic mode performance of EPEE method compared with entropy-based and the patience-based methods. For comparison with the entropy-based method and the patience-based method, their results were plotted as color green and red with one parameter considered to be maximum or minimum. Our method presented superior flexibility over the baselines. }
\label{fig:rep}
\end{figure}

\begin{figure}[h]
    \centering
    \includegraphics[width=0.99\linewidth]{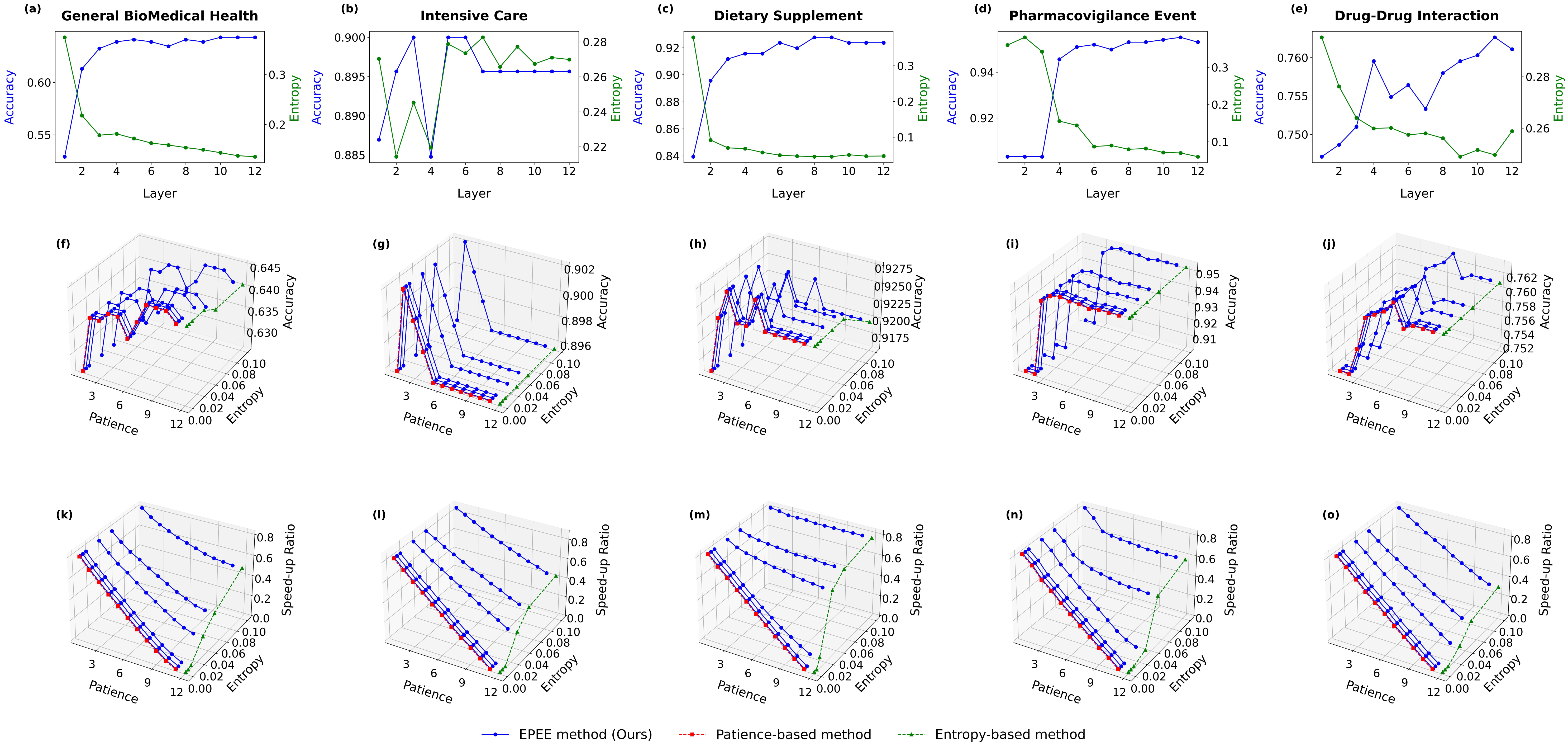}
    \caption{Hyper-parameter analysis with ALBERT on five biomedical text datasets. The results demonstrated the impact of key factors of EPEE on its efficiency and effectiveness.}
    \label{fig:albert}
\end{figure}

\begin{figure}
    \centering
    \includegraphics[width=0.99\linewidth]{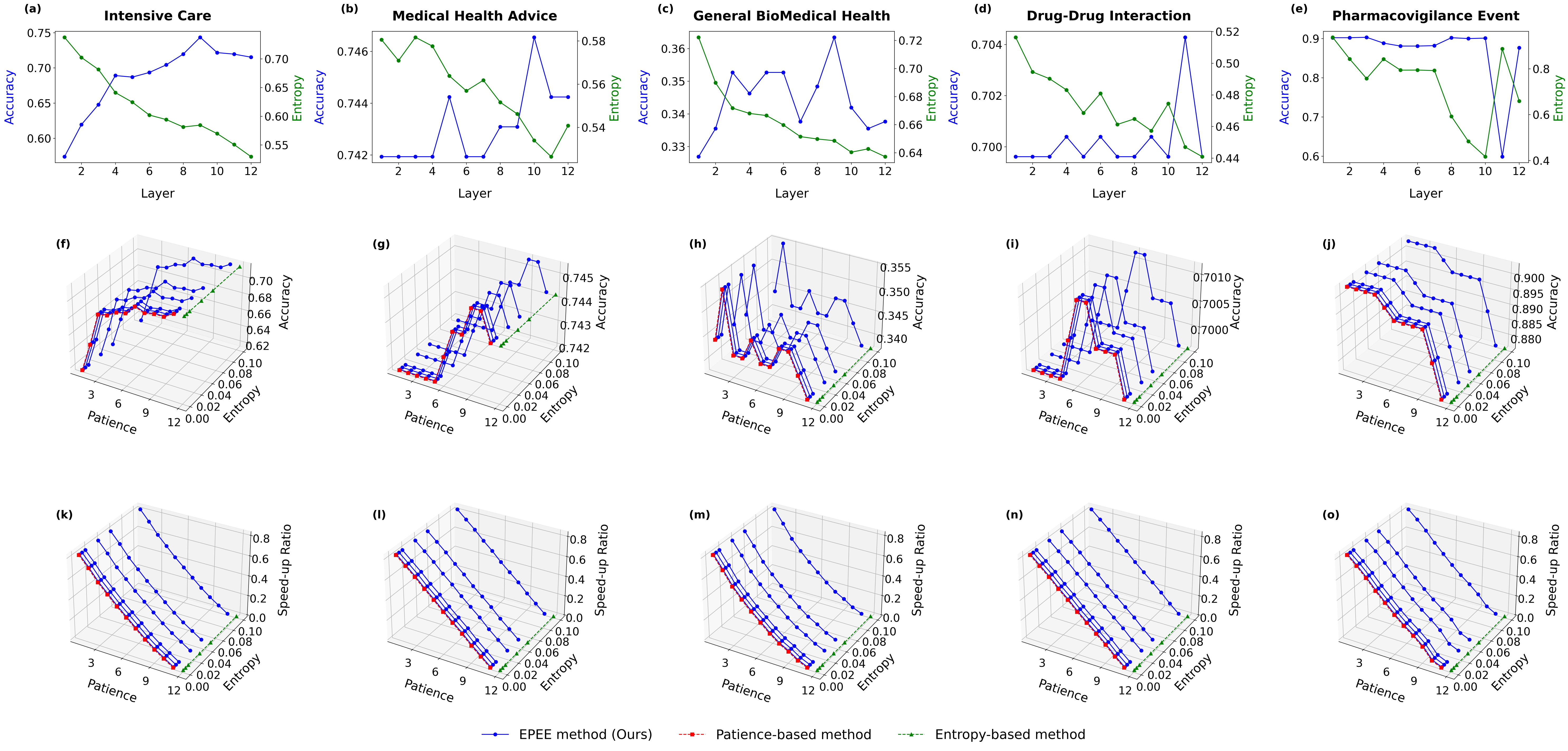}
    \caption{Hyper-parameter analysis with GPT-2 on five biomedical text datasets. The results demonstrated the impact of key factors of EPEE on its efficiency and effectiveness.}
    \label{fig:gpt2}
\end{figure}

\begin{figure}[t]
    \centering
    \includegraphics[width=0.99\linewidth]{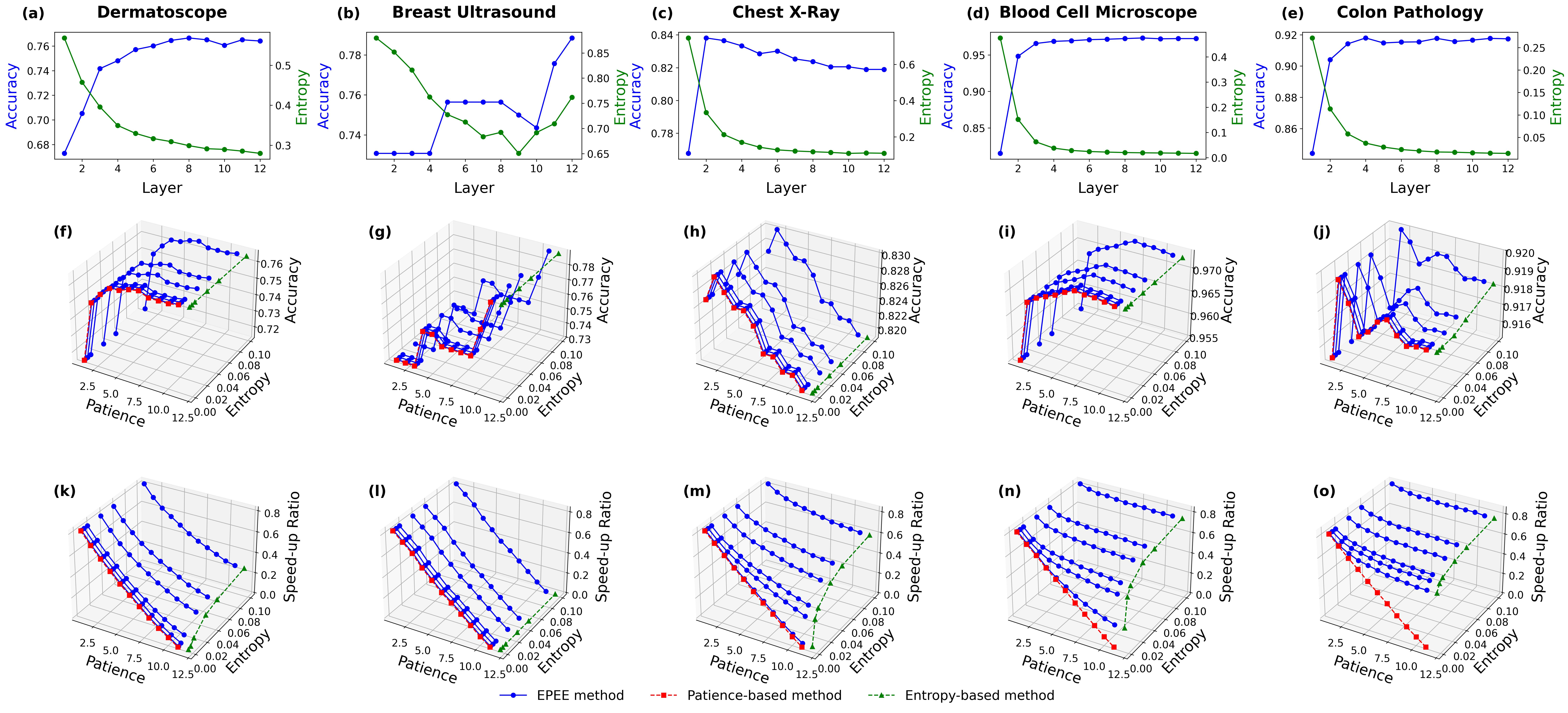}
    \caption{Hyper-parameter analysis with ViT on five medical image datasets. The results demonstrated the impact of key factors of EPEE on its efficiency and effectiveness.}
    \label{fig:cv}
\end{figure}

\section{Results}
In this section, we first demonstrated the efficiency and effectiveness of EPEE on language models under two operational modes: budgeted mode and dynamic mode. Next, we validated the effectiveness of our approach on visual foundation models.
Finally, we showed the proposed EPEE method could be equivalent to the entropy-based method and the patience-based method respectively by changing the hyper-parameters, which demonstrated its superior generalization.


\subsection{Budgeted Mode}
After the LLMs and classifiers were trained jointly, we used each exit (i.e., classifiers after each transformer layer) to perform the classification, relation extraction, and event extraction tasks, and the accuracy and prediction entropy for each layer was shown in Fig. \ref{fig:effe}.
It showed that the prediction entropy decreased with the layer increased, which meant the prediction was more and more confident as the transformer layer increased.
In contrast, the classification accuracy presented an increasing trend as it delved deeper.
The EPEE method worked well with BERT on the MIMIC-ICU dataset. The performance of the first layer was close to the highest layer which was the fourth layer.
Most of the layers outperformed the last layer but the majority of current work uses the last layer directly.
Similarly, for the private dietary supplement use dataset, the accuracy got above 0.8 only after two layers, and it achieved the highest accuracy at the fourth layer while the entropy reached the lowest at the last layer.
In addition, our proposed method EPEE demonstrated similar results for PHEE, DDI, and GIT datasets as they achieved the highest accuracy before the last layer.
The most exciting finding was that the first exit of models trained for Drug review, PHEE, DDI, and Medical health advice datasets exhibited high performance, especially for DDI and PHEE datasets, their first exit performance was close to the highest performance, which indicated huge efficient rewards from a little sacrifice in performance.

To show the robustness of EPEE, hyper-parameters analysis was performed on ALBERT and GPT-2 models. They are both transformer-based models, but ALBERT is the encoder-based and GPT-2 is the decoder-based architecture.
The results were shown in Fig.~\ref{fig:albert}(a-d) and Fig.~\ref{fig:gpt2}(a-d) separately.
We could observe similar results: 1) the best performance may not happen in the last layer; 2) the first few layers perform well with huge efficiency, which shows the overthinking issue is more severe in the GPT-2 model than in the BERT model.

\subsection{Dynamic Mode}

Unlike the budgeted mode, which sets a fixed computational depth for all inputs, the dynamic mode enables adaptive inference depth, allowing models to determine the optimal exit point for each input dynamically. Simpler cases exit earlier to minimize computational overhead, while more complex cases continue processing through deeper layers to ensure accuracy.

To comprehensively evaluate the flexibility of our method, we conducted a grid search over different entropy and patience settings. The results for the BERT model, presented in Fig. \ref{fig:rep}, illustrated the trade-off between accuracy and inference speed across varying configurations. Notably, for the dietary supplement usage classification dataset, the highest accuracy was achieved with a patience value of 2 or 3 across all entropy thresholds. This aligned with the budgeted mode findings, where the fourth layer yielded the highest classification accuracy.

The broader experimental results across multiple datasets reveal two consistent trends: (1) Increasing the patience value generally improves accuracy, as it allows the model to confirm predictions over multiple layers. (2) Higher entropy thresholds and lower patience values result in greater inference speed-up, reducing computational cost but potentially impacting accuracy.

To further validate the robustness of the dynamic mode across different model architectures, we extended the experiments to ALBERT and GPT-2, as shown in Fig. \ref{fig:albert} (e-l) and \ref{fig:gpt2} (e-l). The results confirmed that EPEE maintained its flexibility across transformer encoder and decoder architectures. Notably, for GPT-2, the first few exits already achieved near-peak accuracy, indicating that the overthinking issue was more pronounced in decoder-based architectures. Moreover, the dynamic mode effectively balanced speed and accuracy, demonstrating that even with different backbone models, the optimal exit layer varied based on input complexity.

These findings highlight the adaptability of EPEE in dynamic mode, enabling fine-grained control over computational efficiency and predictive performance.
The dynamic mode allows users to select the best entropy and patience combination to achieve high performance and low latency at the same time.
Compared to existing early exiting strategies, EPEE provides a more flexible mechanism to balance speed and accuracy, making it particularly suitable for real-world biomedical applications where inference latency and decision reliability are critical.

\subsection{EPEE for Biomedical Vision}
To demonstrate the generalizability of our proposed EPEE method beyond language models, we extended our experiments to vision foundation models. Specifically, we evaluated EPEE on ViTs, which have emerged as a powerful architecture for medical image analysis. 

In the budgeted mode, as shown in Fig. \ref{fig:cv} (a-e), we assessed how the accuracy and entropy of predictions evolve across different transformer layers in ViTs. Similar to our findings in language models, we observed that predictions become more confident (i.e., entropy decreases) as layer depth increases. However, classification accuracy tends to stabilize after a few intermediate layers, suggesting that deeper layers may not always be necessary for optimal performance.

For most datasets, high classification accuracy was achieved at early exits, demonstrating that overthinking also exists in vision models. For example, in the PathMNIST dataset, intermediate layers produced comparable accuracy to the final layer, while significantly reducing computational cost. This finding reinforces the importance of early exiting in medical imaging, where efficiency is crucial for real-time diagnosis and clinical decision-making.

In addition, Fig. \ref{fig:cv} (f-o) illustrates how varying entropy and patience thresholds impact the accuracy and speed-up ratio across different datasets. 
The dynamic mode gives users a chance to improve performance and efficiency at the same time by refining the entropy and patience settings.
For example on the BloodMNIST dataset, accuracy could reach 0.97 with the speed-up ratio of 0.7 by setting the entropy threshold to 0.1 and patience to 6.

\begin{figure}[b]
\centering
\begin{subfigure}{0.24\linewidth}
    \centering
    \includegraphics[width=\linewidth]{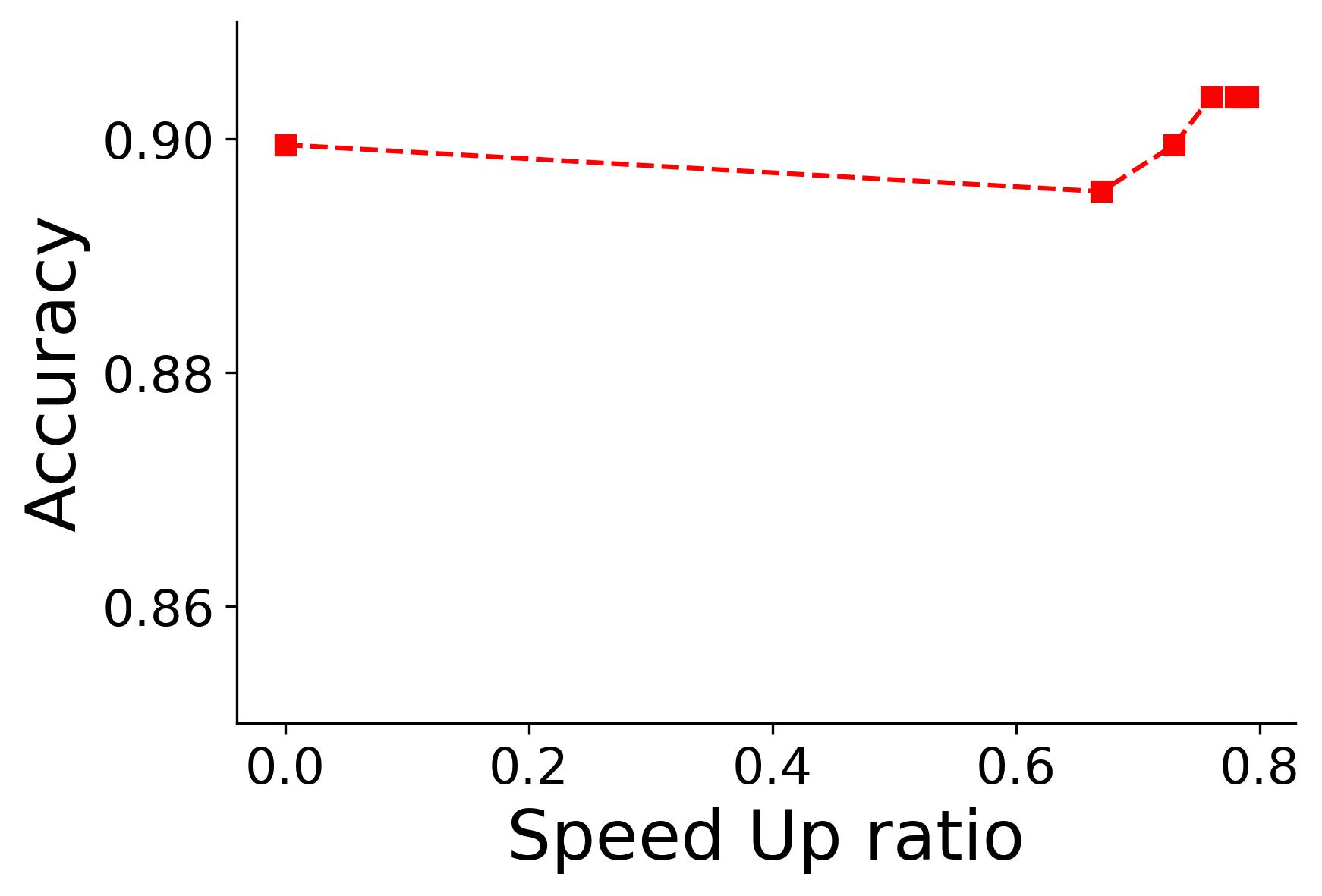}
    \caption{Entropy-based method}
    \label{fig:rep11}
\end{subfigure}
\begin{subfigure}{0.24\linewidth}
    \centering
    \includegraphics[width=\linewidth]{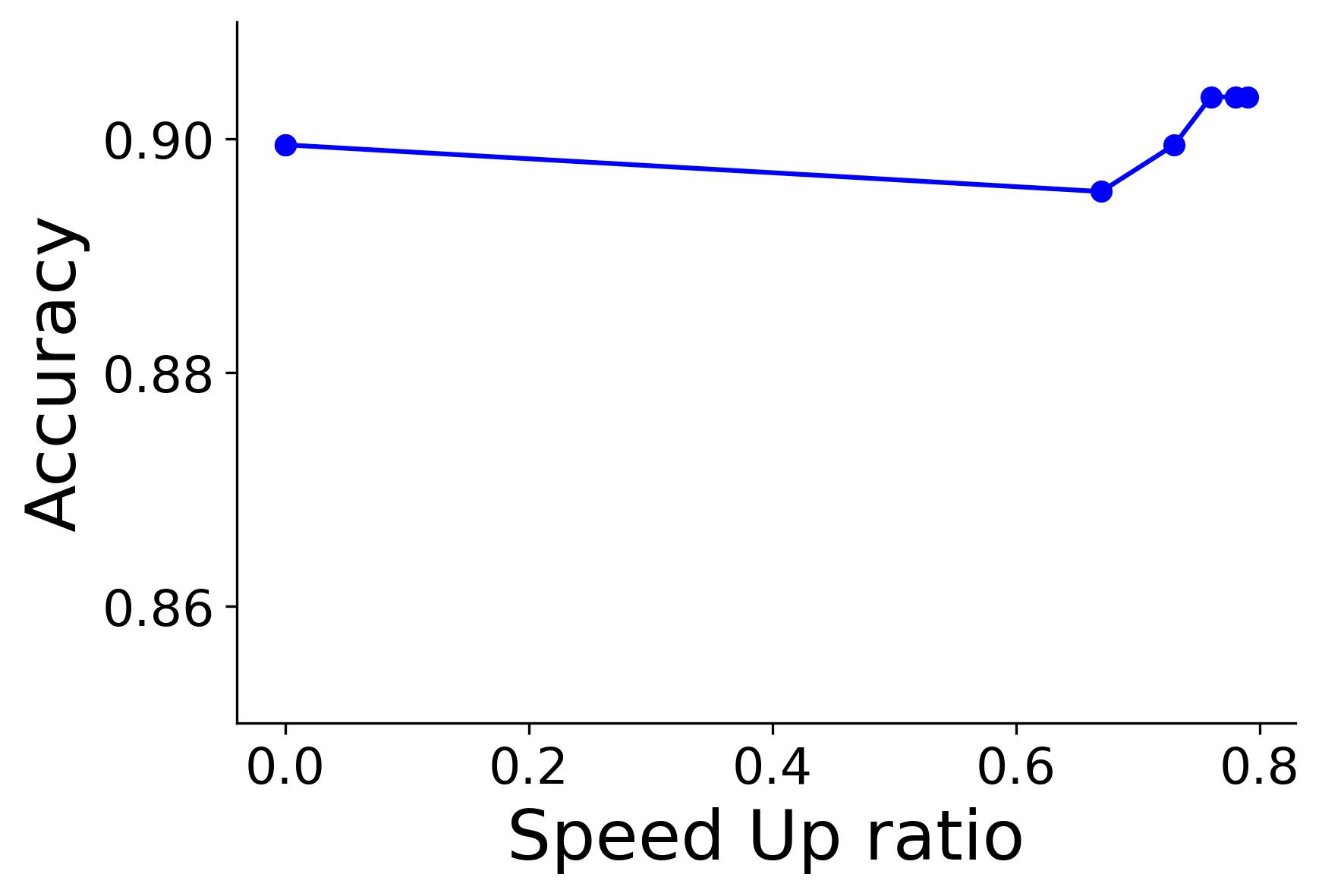}
    \caption{Our method (patience = 12)}
    \label{fig:rep12}
\end{subfigure}
\begin{subfigure}{0.24\linewidth}
    \centering
    \includegraphics[width=\linewidth]{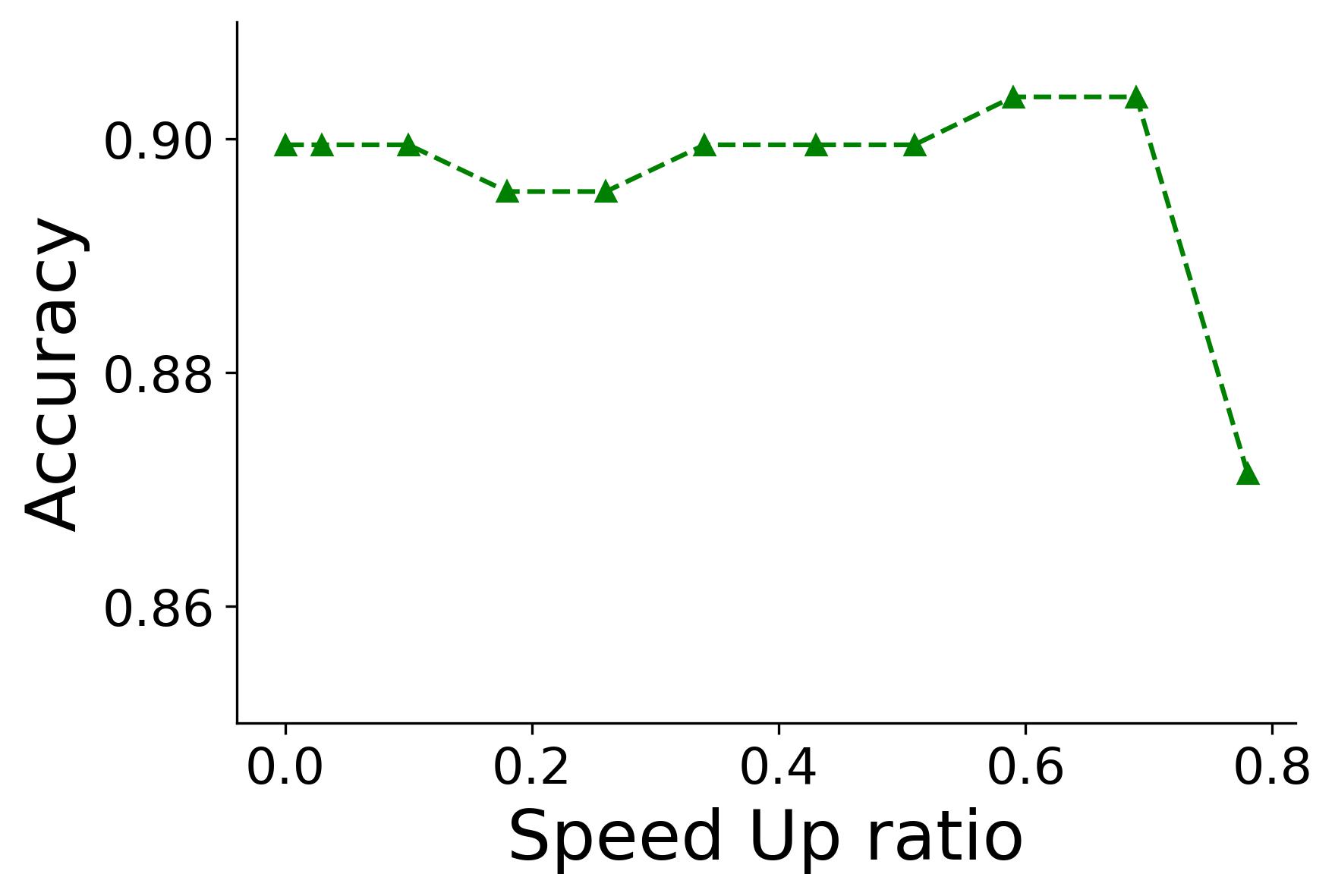}
    \caption{Patience-based method}
    \label{fig:rep21}
\end{subfigure}
\begin{subfigure}{0.24\linewidth}
    \centering
    \includegraphics[width=\linewidth]{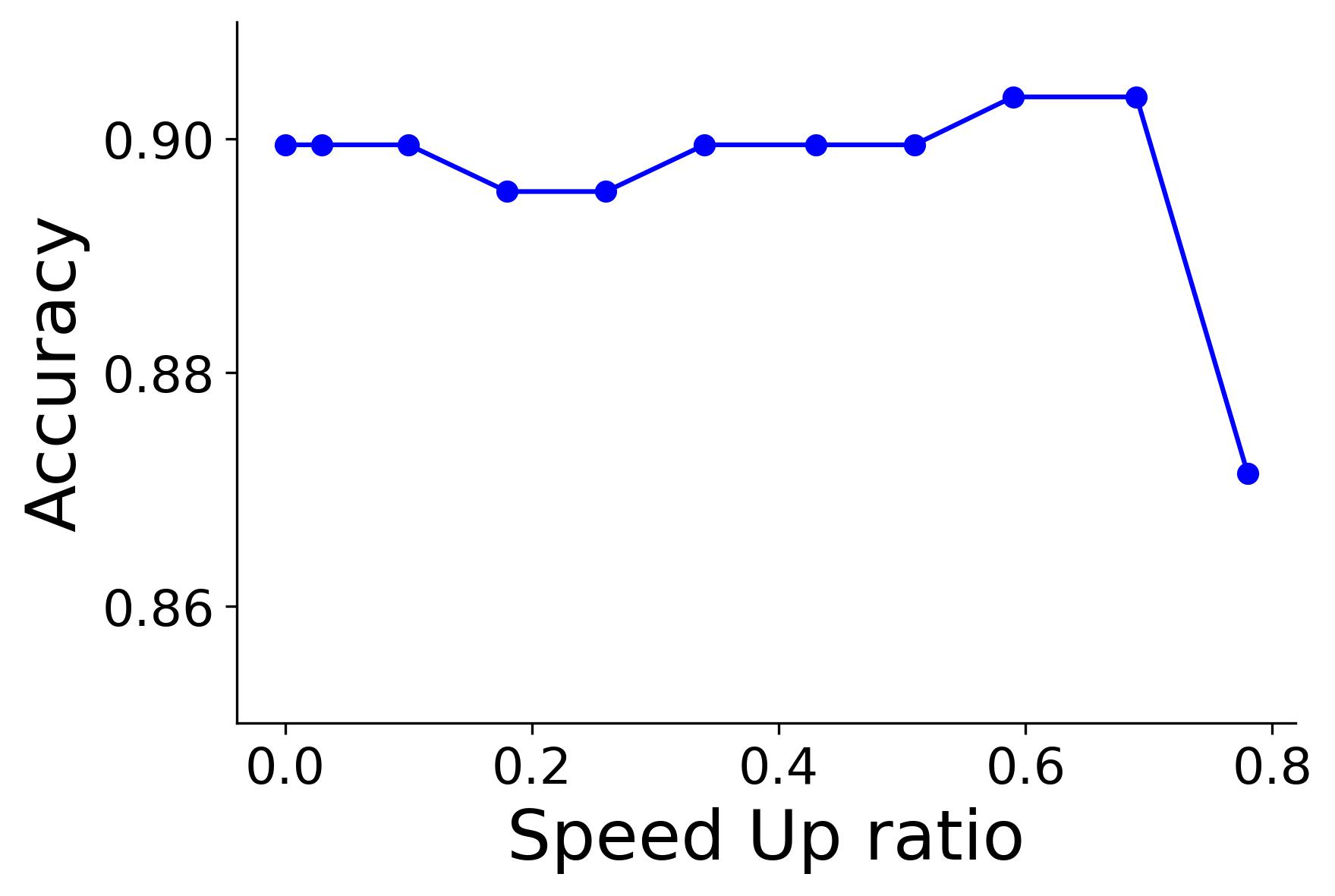}
    \caption{Our method (entropy = 0)}
    \label{fig:rep22}
\end{subfigure}
\caption{Degeneracy analysis of EPEE. Our method could be simplified into the entropy-based or the patience-based methods by invalidating a parameter.}
\label{fig:reduce}
\end{figure}

\subsection{Method Generalization and Special Cases}
To illustrate that the EPEE method not only is flexible but also covers the function of the entropy-based method or the patience-based method, we examine its behavior in specific limiting cases. 
As shown in Fig. \ref{fig:reduce} (a) and (b), if we set the patience to \(M\) (i.e. the number of transformer layers of PLM), then the patience counter would only possibly be satisfied if it reaches to the last layer. 
However, when it reaches the last layer, it predicts the output regardless of whether any criteria are satisfied.
In this setting, the patience criterion loses control of decisions and thus our EPEE method reduces to the entropy-based method.
On the other hand, as shown in Fig. \ref{fig:reduce} (c) and (d), if the entropy threshold is set to 0, the entropy criterion is prohibited because the entropy criterion would never be satisfied. Then the patience criterion would be the only way to early exit.
Therefore, the EPEE method is more general, and the two previous methods are the special cases of the EPEE method.

\section{Discussion}
In this study, we introduced the EPEE method, specifically designed to address accuracy and latency challenges in the biomedical and healthcare domains. We evaluated EPEE across three tasks—classification, relation extraction, and event extraction—using twelve diverse datasets: MIMIC-ICU, dietary supplement usage, drug review, PHEE, DDI, GIT, and medical health advice, pathMNIST, PneumoniaMNIST, BloodMNIST, DermaMNIST, BreastMNIST. The method was evaluated against two existing approaches, entropy-based and patience-based early exiting, and was implemented in three main structures: transformer encoder, transformer decoder, and vision transformer via four pre-trained models: BERT, ALBERT, GPT-2, and ViT under two operational modes: budgeted and dynamic. Our findings highlight both the necessity of early exiting in this domain and the advantages offered by our EPEE method in improving both performance and computational efficiency.

The budgeted mode results clearly emphasize the importance of early exiting in the biomedical domain. Traditional inference approaches, which rely on the classifier in the final layer of the model, often lead to the "overthinking" issue—where additional processing in deeper layers does not contribute to better predictions and may even degrade performance. This phenomenon was evident in all models we used- BERT, ALBERT, GPT-2, and ViT, where exiting earlier at intermediate layers resulted in better performance compared to utilizing the final layer. The severity of this overthinking issue is expected to increase with larger models, as all the first exits in the GPT-2 model achieved high accuracy, making early exiting an essential strategy for future model deployments.

Beyond improving performance, early exiting also delivers significant efficiency gains in inference time. For example, in the budgeted mode, we observed that the fourth exit of the BERT model trained on the dietary supplement usage dataset achieved the highest F1 score while consuming only one-third of the inference time required by the full model. This finding underscores the practical value of early exiting, particularly in scenarios such as high-traffic healthcare environments, where both performance and response times are critical.

Interestingly, our experiments revealed consistent patterns across datasets. For each dataset, there is typically a specific transformer layer or a small range of transitional layers (2-3 layers) after which the performance stabilizes and closely approximates that of the final layer.
For instance on the BERT model, in the Drug Review, PHEE, GIT, and Medical Health Advice datasets, models achieved near-optimal performance around the 9th transformer layer.
These layers may have already captured most of the key features, making further computations in deeper layers potentially redundant. This trend is consistently observed across different tasks and datasets, indicating that the computational demand of the model does not grow linearly but instead exhibits diminishing returns beyond a certain depth, which illustrates the importance of the early exiting method.

In the dynamic mode, our proposed EPEE method demonstrated exceptional flexibility and adaptability compared to the entropy-based and patience-based methods. While the latter two approaches are governed by a single parameter, EPEE leverages a combination of entropy and patience thresholds to provide a finer level of control over the speed-up ratio. As illustrated in Fig. \ref{fig:rep} and Fig. \ref{fig:albert}, the EPEE method allows users to dynamically adjust the trade-off between accuracy and inference speed by tuning these thresholds. This capability is particularly valuable in real-world applications, where different tasks or operational environments may demand varying levels of precision and computational efficiency.

A key advantage of EPEE in dynamic mode is that it enables the discovery of optimal configurations where models achieve both high accuracy and significant speed-up, striking a balance that neither entropy-based nor patience-based methods can achieve individually. Specifically, our experiments reveal that by carefully selecting the entropy and patience parameters, EPEE consistently identifies configurations where inference is accelerated while maintaining performance comparable to full-depth execution. On multiple datasets, we observe speed-up gains and performance gains could be achieved at the same time.

In addition, through grid-search optimization of entropy and patience thresholds, the EPEE method can achieve any desired speed-up ratio to satisfy user needs. This flexibility enables practitioners to explore a wider range of trade-off configurations, allowing for better alignment with specific application requirements. Notably, our method addresses a key limitation of entropy-based methods, where the speed-up ratio often exhibits abrupt changes over a narrow parameter range, making fine-tuning challenging. By pairing specific entropy values with corresponding patience values, EPEE provides a smoother, more controlled adjustment, ensuring both efficiency and reliability in the model's performance.


The experiments on medical computer vision reveal the robustness of the EPEE method for different architectures, including transformer encoder, decoder, and vision transformer, which guarantees that the EPEE method is compatible with all foundation models.
In the medical computer vision domain, overthinking and latency issues also exist, as shown in Fig. \ref{fig:cv}.
The ViT can always provide higher or comparable accuracy with a better speed-up ratio, which lays the ground for efficient and accurate vision foundation models in biomedical and healthcare domains.

The final notable strength of the EPEE method lies in its generality and inclusivity. By appropriately setting one of its parameters to an extreme value, the EPEE framework can effectively reduce to either the entropy-based or patience-based methods, demonstrating its generalization on covering main existing approaches. This capability highlights the versatility of EPEE, as it not only enhances flexibility but also integrates the strengths of prior methods under a unified framework.


The findings of this study underscored the critical role of early exiting strategies in optimizing the performance and efficiency of foundation models in the biomedical and healthcare domains. The results suggested that incorporating the advanced EPEE method could address the overthinking issue inherent in deep transformer architectures while enabling models to operate effectively within the constraints of real-world applications. 
Future work could explore extending the EPEE framework to even larger and more diverse datasets, as well as investigating its applicability to other domains beyond healthcare. Additionally, integrating EPEE with emerging multi-modal foundation models that process both text and image inputs simultaneously, could further reveal its capabilities and widen its scope of impact.


\section{Methods}
\subsection{Preliminaries}
In this subsection, we present the essential background and define the mathematical notations for the early exiting method. This study focuses on a multi-class classification setting, where the dataset is denoted as \((X, Y)\), with individual samples represented by \((x_i, y_i)\) for \(i = 1, 2, \ldots, N\). Here, \(x_i \in X\) represents the input sentence, and \(y_i \in Y\) corresponds to its associated label. 
The classification task involves a class space denoted by \(K\). We define \(M\) as the total number of Transformer layers, \(d\) as the hidden layer dimension, and \(s_m\) as the hidden state obtained after the \(m\)-th layer, where \(m \in \{1, 2, \ldots, M\}\).

\subsubsection{Entropy-based Early-Exiting Method}
As illustrated in Figure~\ref{fig:methods}, early exiting architectures incorporate exit points at each Transformer layer. For a model with \(M\) Transformer layers, \(M\) classifiers \(f_m(s_m; \theta_m): s_m \rightarrow K\) (\(m = 1, 2, \ldots, M\)) are designated at these layers. Each classifier maps the hidden state \(s_m\) of its respective layer to a probability distribution \(p_m(s_m; \theta_m)\) over \(|K|\) classes using the softmax function.
The confidence level of each layer \(m\) is quantified using the entropy of the predicted class distribution \(p_m\). Normalized entropy, which serves as a measure of confidence, is calculated as follows:
\begin{equation}
H_m = -\frac{\sum_{k=1}^{|K|} p_m^k \log p_m^k}{\log |K|},
\end{equation}
where \(p_m^k\) represents the probability assigned to the \(k\)-th class by the \(m\)-th Transformer layer. A lower entropy value \(H_m\) signifies higher confidence in the prediction. If \(H_m\) is less than a predefined threshold \(\tau\), the prediction at layer \(m\) is considered confident. Otherwise, the process proceeds to the next Transformer layer.
If none of the classifiers generate a confident prediction, the final classifier at the last layer outputs the prediction, irrespective of its confidence level.

\subsubsection{Patience-based Early-Exiting Method}
As illustrated in Figure~\ref{fig:methods}, a patience counter \(P\) is maintained to track how many consecutive classifiers predict the same class. If two consecutive predictions differ, the patience counter is reset to 1. At each layer \(m\), the patience counter is updated as follows:
\begin{equation}
P_m = 
\begin{cases} 
P_{m-1} + 1, & \text{if } \arg \max_k p_m^k = \arg \max_k p_{m-1}^k, \\ 
1, & \text{otherwise}.
\end{cases}
\end{equation}
When \(P_m\) reaches a predefined threshold \(P_t\) (the patience parameter), the model exits early at layer \(m\). If this condition is never satisfied, the final classifier at the last layer \(M\) produces the prediction. This approach allows the model to exit early when multiple classifiers consistently predict the same result, ensuring high confidence in the prediction.


\subsection{EPEE Method}
The entropy-based acceleration method has garnered significant attention due to its simplicity, efficiency, and flexibility. However, it suffers from reduced accuracy when the entropy threshold is increased to accelerate inference. Conversely, the patience-based acceleration method achieves state-of-the-art performance but encounters inefficiencies: it tends to over-process simple inputs when the patience parameter is set too high, while under-processing certain inputs when the parameter is set too low. Furthermore, both methods rely on a single parameter to control the speed-up ratio, which can be inconvenient when aiming for a specific budgeted speed-up ratio. Consequently, there is a pressing need for a novel method to address these limitations.

To bridge this gap, we propose a novel early exiting method, termed EPEE (Entropy- and Patience-based Early Exiting), as illustrated in Fig.~\ref{fig:methods}. Similar to the existing approaches, our method evaluates whether to exit at each attention layer during inference. However, EPEE simultaneously employs two exiting criteria, thereby inheriting the strengths of both methods. 

For an input sentence $x$, at the \(m\)-th layer, the model exits directly if the entropy score is below a predefined threshold or if the patience counter reaches a predefined value. Otherwise, the patience counter is updated: it increments by 1 if the current layer's prediction matches that of the previous layer; otherwise, it resets to 1. If no stopping criterion is met, the classifier at the final transformer layer generates the prediction. The mechanism is summarized as:

\begin{equation}
\text{Decision}_m = 
\begin{cases} 
\text{exit}, & \text{if } H_m < \tau \quad \text{or} \quad P_m = P_t, \\ 
P_m = P_{m-1} + 1, & H_m \geq \tau \quad \text{and} \quad \arg\max_k p_m^k = \arg\max_k p_{m-1}^k, \\ 
P_m = 1, & \text{otherwise}.
\end{cases}
\end{equation}

The primary advantage of the proposed EPEE method lies in its flexibility. While the entropy-based and patience-based methods rely solely on entropy and patience counters, respectively, to determine the exit, EPEE leverages both, enabling a more versatile control over the speed-up ratio. Moreover, EPEE encompasses both existing methods as special cases: setting the entropy threshold to 0 reduces EPEE to the patience-based method, while defining the patience parameter as \(M\) (i.e., all layers) reduces EPEE to the entropy-based method.

Additionally, EPEE resolves the limitations of the individual methods by effectively combining their strengths. The entropy-based approach efficiently handles simple input sentences with high confidence but may falter with complex inputs. In contrast, the patience-based method can waste computational resources on simple inputs due to its fixed patience threshold. By adopting a small entropy threshold, EPEE ensures that simple sentences exit quickly with high confidence, while more complex inputs utilize the patience counter to exit appropriately. This ensures that complex sentences are processed efficiently without waiting until the final layer, while simple inputs are resolved expeditiously.

\subsection{Study Design}

\subsubsection{Datasets}
We evaluated EPEE against alternative methods using seven biomedical text datasets across three core tasks: Classification (3 datasets), Relation Extraction (2 datasets), and Event Extraction (1 dataset). Additionally, we conducted experiments on five medical image classification datasets.

Among the selected datasets, the Dietary Supplements Usage Status dataset~\cite{fan2018using, fan2017classifying, ramie} is a private dataset developed using clinical notes from the University of Minnesota. This dataset specifically targets mentions of dietary supplements, comprising a total of 3,000 annotated sentences categorized into four use status classes: Continuing (C), Discontinued (D), Started (S), and Uncertain (U). The dataset captures mentions of dietary supplements frequently used by patients in clinical settings. The 25 dietary supplements included in the dataset are: \textit{Alfalfa, Biotin, Black Cohosh, Coenzyme Q10, Cranberry, Dandelion, Echinacea, Fish Oil, Flax Seed, Folic Acid, Garlic, Ginger, Ginkgo, Ginseng, Glucosamine, Glutamine, Kava Kava, Lecithin, Melatonin, Milk Thistle, Saw Palmetto, St. John's Wort, Turmeric, Valerian, Vitamin E}.
All other datasets used in this study are publicly available. A summary of the datasets is presented in Table~\ref{tab:dataset_overview}.




\begin{table}[h]
\centering
\resizebox{\textwidth}{!}{%
\begin{tabular}{cccccccccc}
\toprule
\textbf{Dataset} & \textbf{Data Status} & \textbf{Data Source} & \textbf{Data Type} & \textbf{Task} & \textbf{Train} & \textbf{Dev} & \textbf{Test} & \textbf{Classes}  \\
\midrule
MIMIC-III~\cite{hager2024evaluation} & Public & Beth Israel Deaconess Medical Center & Intensive Care Unit Record & Classification & 3,861 & 483 & 483 & 4 \\
Dietary Supplement Usage~\cite{fan2017classifying} & Private & University of Minnesota & Electrical Health Record & Classification  & 2,000  & 230  & 230  & 4    \\
Drug Review~\cite{Grer2018AspectBasedSA} & Public & Patient Reviews & Patient-generated Text & Classification  & 161,297 & 53,766 & 53,766 & 2    \\
Medical Health Advice~\cite{yu2019detecting} & Public & PubMed & Medical Literature & Classification   & 6,940  & 868  & 868  & 3    \\
DDI~\cite{segura2013semeval} & Public &  Medline Abstract & Medical Literature & Relation Extraction  & 11,556 & 1,285 & 3,020 & 5    \\
GIT~\cite{li2023petailor} & Public & PubMed Abstract & Medical Literature & Relation Extraction   & 3,734  & 465  & 492  & 22  \\
PHEE~\cite{sun2022phee} & Public & Literature and Reports & Medical Literature & Event Extraction & 2,898  & 961  & 968  & 2   \\
\hline
PathMNIST~\cite{kather2019predicting,yang2023medmnist} & Public & NCT Biobank and the UMM Pathology Archive & Colon Pathology & Classification & 89,996 & 10,004 & 7,180  & 9   \\
PneumoniaMNIST~\cite{kermany2018identifying,yang2023medmnist} & Public &	Children Hospital & Chest X-Ray & Classification	& 4,708 & 524 & 624 & 2\\
BreastMNIST~\cite{al2020dataset,yang2023medmnist} & Public &	Baheya Hospital & Breast Ultrasound & Classification & 546 & 78 & 156 & 2 \\
BloodMNIST~\cite{acevedo2020dataset,yang2023medmnist} & Public & Hospital Clinic of Barcelona & Blood Cell Microscope & Classification & 11,959 & 1,712 & 3,421 & 8 \\
DermaMNIST~\cite{tschandl2018ham10000,codella2019skin,yang2023medmnist} & Public & Medical University of Vienna and Cliff Rosendahl & Dermatoscope & Classification & 7,007 & 1,003 & 2,005 & 7\\
\bottomrule
\end{tabular}
}
\caption{Overview of the data statistics.}
\label{tab:dataset_overview}
\end{table}

\subsubsection{Backbone Models}
Considering the efficiency and widespread impact, this study mainly used BERT~\cite{devlin-etal-2019-bert},  ALBERT~\cite{albert}, GPT-2~\cite{radford2019language} and ViT~\cite{dosovitskiy2021an} as the backbone models.
They consist of three different structures: Transformer encoder~\cite{Vaswani2017attention}, Transformer decoder~\cite{Vaswani2017attention}, and Vision transformer~\cite{Alex2012cnn} where these structures cover all the current foundational models.

\begin{table}[ht]
\centering
\begin{tabular}{ccccccccc}
\toprule
\textbf{Model} & \textbf{Parameter} & \textbf{Backbone} & \textbf{Pre-trained Data} \\
\midrule
BERT~\cite{} & 109M & Transformer encoder & Wikipedia + BooksCorpus  \\
ALBERT & 12M & Transformer encoder & Wikipedia + BooksCorpus  \\
GPT-2 & 124M & Transformer decoder & Wikipedia + News + Books  \\
ViT & 86M & CNN + Transformer encoder & ImageNet~\cite{Deng2009ImageNet}+ LAION~\cite{schuhmann2022laion} + JFT~\cite{zhai2022scaling}\\
\bottomrule
\end{tabular}
\caption{Overview of the foundational models.}
\label{tab:model_overview}
\end{table}

\subsubsection{Training}
During training, all exiting classifiers are jointly optimized using a weighted cross-entropy loss function. Following previous works~\cite{huang2017multi,zhou2020bert,zhang2022pcee}, the loss function is formulated as a weighted average of the cross-entropy losses, given by:
\begin{equation}
\mathcal{L} = \frac{\sum_{m=1}^{M} w_m \cdot H_m(y,p_m)}{\sum_{m=1}^{M} w_m}
\end{equation}
Here, \(H_m(y,p_m) = - \sum_{i=1}^{N} y_i \log ({p_m}_i)\) denotes the cross-entropy loss for the \(m\)-th exit, where \(y\) represents the ground truth, \(p_m\) is the predicted probability distribution at the \(m\)-th exit, and \(N\) is the number of classes. The weight \(w_m\) corresponds to the relative inference cost associated with the \(m\)-th exit.

\subsubsection{Speed-up Ratio}\label{speedup}
The efficiency of the early exiting method is quantified using the speed-up ratio~\cite{zhou2020bert,zhang2022pcee}, which is computed as follows. Let \(M\) denote the total number of layers in the backbone model, and for each test sample \(x_i\) (where \(i = 1, 2, \dots, N\)), let \(\mathbbm{1}_{m_i}\) indicate whether the \(m\)-th transformer layer is utilized during inference for input \(x_i\). The average speed-up ratio over the test set is defined as:
\begin{equation}
\text{Speedup} = 1 - \frac{\sum_{i=1}^{N}\sum_{m_i=1}^{M} \mathbbm{1}_{m_i}}{N \times M}
\end{equation}
This metric is chosen for its linear relationship with actual computational cost. Based on our experiments, it demonstrates a strong correlation with wall-clock runtime while maintaining stability across runs, even in the presence of potential randomness introduced by other processes on the same machine.




\subsubsection{Inference Modes}
During inference, the model with multiple exits can employ two early exiting strategies based on whether the computational budget is predefined.

\noindent
\textbf{Budgeted Exiting} If the computational budget is set, a specific exit \(m^*\) can be chosen, where \(m^*\)-th exiting classifier is used to predict all queries.

\noindent
\textbf{Dynamic Exiting} In this mode, after receiving an input \(x\), the model sequentially predicts using classifiers from beginning to end, reusing computations where feasible. The process continues until an exit criterion is met at layer \(m^* < M\) or the final exit \(M\) is reached. The final prediction combines the current and previous predictions, allowing different samples to exit at varying layers.

\subsubsection{Experimental Settings}
The foundational models and classifier at every layer were trained once and we reloaded the coefficients for different early exiting strategies.
During training, a grid search was conducted to determine optimal hyper-parameters, with batch sizes set at 16, 32, and 128, and learning rates tested at 1e-5, 2e-5, 3e-5, and 5e-5 using the Adam optimizer for 15 epochs. 
All implementations were built on Hugging Face’s Transformers library \cite{wolf2020transformers}, and experiments were conducted on a single Nvidia A100 GPU with 40GB memory.

For inference, we adopted a per-instance approach, setting the batch size to 1 to simulate real-world usage, where individual requests may come from different users at varying times.



\section{Data Availability}
Eleven datasets involved in this study are publicly available from the following links: 

Drug Review: \href{https://archive.ics.uci.edu/dataset/462/drug+review+dataset+drugs+com}{https://archive.ics.uci.edu/dataset/462/drug+review+dataset+drugs+com}

Medical Health Advice: \href{https://huggingface.co/datasets/medalpaca/medical_meadow_health_advice}{https://huggingface.co/datasets/medalpaca/medical\_meadow\_health\_advice}

DDI: \href{https://github.com/isegura/DDICorpus}{https://github.com/isegura/DDICorpus}

GIT: \href{https://github.com/ToneLi/BIoMedRAG/tree/main/dataset/0_GM-CIHT}{https://github.com/ToneLi/BIoMedRAG/tree/main/dataset/0\_GM-CIHT}

PHEE: \href{https://github.com/zhaoyuesun/phee}{https://github.com/zhaoyuesun/phee}

PathMNIST, PneumoniaMNIST, DermaMNIST, BloodMNIST and BreastMNIST: \href{https://medmnist.com/}{https://medmnist.com/}

\section{Code Availability}
The code is publicly available on Github: \href{https://github.com/Learner4everrr/EPEE}{https://github.com/Learner4everrr/EPEE}

\section{Acknowledgements}
This work was supported by the National Institutes of Health’s National Center for Complementary and Integrative Health under grant numbers R01AT009457 and U01AT012871, the National Institute on Aging under grant number R01AG078154, the National Cancer Institute under grant number R01CA287413, the National Institute of Diabetes and Digestive and Kidney Diseases under grant number R01DK115629, and the National Institute on Minority Health and Health Disparities under grant number 1R21MD019134-01.
Many thanks to Yuqian Chen for her help with the drawing.

\section{Author Contributions}
Z.Z. and R.Z. designed the study. 
Z.Z. and S.Z. performed the data collection.
Z.Z. implemented the code, and conducted experiments. 
Z.Z. and S.Z. drafted the manuscript. 
R.Z. supervised the study. 
All authors contributed to the research discussion, manuscript revision, and approval of the manuscript for submission. 

\section{Competing Interests}
The authors declare no competing interests.

\bibliography{0_main}

\end{document}